**Authors**
Hend Al-Khalifa*
Malak Mashaabi
Ghadi Al-Yahya
Raghad Alnashwan

**Affiliations**
Information Technology Department, College of Computer and Information Sciences, King Saud University

**Corresponding author's email address and Twitter handle**
hendk@ksu.edu.sa
twitter: @hend_alkhalifa

**Keywords**
Privacy policy, Personal Data Protection Law, Saudi websites.

**Abstract**
This paper introduces the Saudi Privacy Policy Dataset, a diverse compilation of Arabic privacy policies from various sectors in Saudi Arabia, annotated according to the 10 principles of the Personal Data Protection Law (PDPL); the PDPL was established to be compatible with General Data Protection Regulation (GDPR); one of the most comprehensive data regulations worldwide. Data were collected from multiple sources, including the Saudi Central Bank, the Saudi Arabia National United Platform, the Council of Health Insurance, and general websites using Google and Wikipedia. The final dataset includes 1,000 websites belonging to 7 sectors, 4,638 lines of text, 775,370 tokens, and a corpus size of 8,353 KB.
The annotated dataset offers significant reuse potential for assessing privacy policy compliance, benchmarking privacy practices across industries, and developing automated tools for monitoring adherence to data protection regulations. By providing a comprehensive and annotated dataset of privacy policies, this paper aims to facilitate further research and development in the areas of privacy policy analysis, natural language processing, and machine learning applications related to privacy and data protection, while also serving as an essential resource for researchers, policymakers, and industry professionals interested in understanding and promoting compliance with privacy regulations in Saudi Arabia.

## Specifications table

| Subject | Data Science |
|---|---|
| Specific subject area | Analysis, Text Classification |
| Type of data | Table (CSV file) |
| How the data were acquired | Data were acquired from various sources, including the Saudi Central Bank, Saudi Arabia National United Platform, and the Council of Health Insurance. General websites were found using Google and Wikipedia, with classification-based searching on Wikipedia for easier discovery of Saudi-owned websites. Due to irrelevant information and other challenges, data were manually scraped, extracting only required information. Only the first web page of privacy policies was collected, discarding PDF files, additional links, and websites with disabled copy-text functionality. |
| Data format | Raw, Annotated |
| Description of data collection | Data collection for privacy policies involved multiple phases. In the first phase, text was manually extracted while scraping, omitting introductions, contact details, and hyperlinks. In the second phase, policies were combined into a CSV file. During the third phase, data normalization was conducted, and Python regular expressions were used to remove Arabic diacritics, English letters, symbols, and numbers. In the final phase, blank lines, rows, and multi spaces in cells were eliminated, resulting in a cleaned dataset. |
| Data source location | Country: Saudi Arabia |
| Data accessibility | Repository name: Github<br><br>Direct URL to data: https://github.com/iwan-rg/Saudi_Privacy_policy |
| **Related research article** | N/A |

**Value of the data**

- The dataset provides a comprehensive and diverse collection of privacy policies across various sectors in Saudi Arabia, enabling researchers to study privacy practices and their alignment with the Personal Data Protection Law.
- Researchers, policymakers, and industry professionals focused on privacy and data protection in Saudi Arabia can benefit from these data.
- The dataset can be used to assess privacy policy compliance, benchmark privacy practices across industries, and develop automated tools for monitoring adherence to data protection regulations.
- The annotated nature of the dataset allows for the development and evaluation of machine learning models and natural language processing applications related to privacy policy analysis.
- The dataset can facilitate comparative studies on privacy policies between Saudi Arabia and other countries or regions, providing insights into cross-cultural and cross-sectoral privacy practices.
- The Saudi Privacy Policy Dataset can be used as a baseline for tracking the evolution of privacy policies over time, enabling the identification of trends and potential areas for improvement in data protection.

**Objective**

The Saudi Privacy Policy Dataset was generated to address the lack of comprehensive and annotated privacy policy datasets specifically tailored to the Saudi Arabian context. As privacy and data protection become increasingly important issues worldwide, understanding the alignment of privacy practices with the Personal Data Protection Law (PDPL) in Saudi Arabia is crucial. By gathering privacy policies from various sectors and annotating them according to the PDPL's 10 principles, this dataset provides a valuable resource for researchers, policymakers, and industry professionals interested in privacy and data protection in the region. The dataset aims to facilitate further research and development in privacy policy analysis, natural language processing, and machine learning applications related to privacy and data protection. Moreover, it enables comparative studies, benchmarking, and compliance assessments, contributing to the growing body of research on privacy and data protection, particularly within the context of Saudi Arabia.

## Data description

**Table 1**
Category Description and number of websites for each category

| Category | Description | File Number |
|---|---|---|
| Finance | Contains Banks and anything related to finance | 52 |
| Government | Contains Governments and anything related to this aspect with (gov) extension | 80 |
| Healthcare | Contains Hospitals, Charites, Clinics, etc. | 27 |
| Educational | Contains Collages, Institutes, Academes, etc. | 42 |
| News | Contains News, Magazines, and Articles, etc. | 19 |
| E-commerce | Contains anything related to commercial transactions | 625 |
| Other | Contains any website that does not belong to the remaining categories | 155 |
| **Total** | | **1000** |

The dataset was collected on December 2022. The total number of privacy policies files collected was 1036. However, due to certain websites' downtime, and others' incorrectly stated privacy policies, we had to exclude some privacy policy websites. The final number of the extracted privacy policies was 1000 files. Table 1 illustrates the category description and the number of webpages collected for each category.

The final dataset was saved in CSV file consisting of 4638 lines of text. Each line is labeled with an appropriate PDPL category that is defined as shown in Figure 1. The number of tokens is 775,370 and the size of the dataset equals to 8,353 KB.

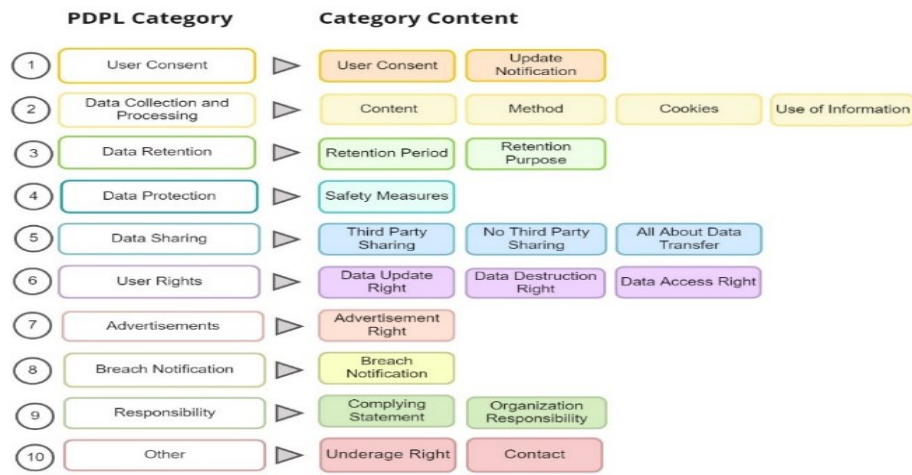

**Fig. 1.** PDPL Annotation Category

## Experimental design, materials and methods

The Personal Data Protection Law (PDPL) was published by Saudi Arabia on 24th September 2021 and came into effect on 23rd March 2022 [1]. Established to be compatible with the GDPR [2], the PDPL applies to entities processing the personal data of Saudi residents [3]. The main principles of the PDPL include: User Consent, Data Collection and Processing, Data Retention,

Data Protection, Data Sharing, Data Rectification, Data Destruction, Data Access, Advertisements, Breach Notification, and Responsibility [1].

In the process of preparing the Saudi Privacy Policy Dataset and its annotations aligned with the Personal Data Protection Law principles, we followed common steps, which included data acquisition from various sources, manual data scraping to extract required information, annotation of the dataset according to the 10 principles of the Personal Data Protection Law, and finally, data cleaning and preprocessing.

### 1) Data Acquisition

We gathered our data from a variety of sources, including the Saudi Central Bank[1] for websites related to finance, the Saudi Arabia National United Platform[2] for websites related to government, and the Council of Health Insurance[3] for websites related to healthcare. To find general websites in Saudi Arabia, we also used Google and Wikipedia. The use of searching by classification on Wikipedia helped us find websites and companies owned by Saudi Arabia more easily. While choosing the privacy policy websites, we found that there is much irrelevant information on each webpage. Therefore, we decided to extract the required information manually. We considered collecting only the first webpage of the privacy policy on each website and discarded the following: any privacy policy written in a PDF file, any links in websites that indicate more privacy policy content, and any website where copy-text functionality was disabled.

### 2) Data annotation

The PDPL principles have been summarized into 10 categories numbered from 1-10 for the annotation purpose. The explanation of the Annotation category is shown in Fig. 1. Each category represents a PDPL principle we defined in our summary, However, category number 10 represents principles mentioned on the collected websites but are not included in our summary, such as the contact information and the underage rights. Right next to the PDPL categories are the category content specified for each category. We used Microsoft Excel as our annotation tool to annotate the policy text with the policy number assigned to it and save it as a CSV file format. The annotation procedure was carried out by the authors, who hold Bachelor's degrees in the field of Information Technology, with a native Arabic language background, and a strong understanding of the privacy policies. Each annotator annotated 333 policies by assigning the appropriate label to each sentence/paragraph.

The distribution of work was carried out among the annotators in such a way that each one was assigned specific websites to annotate. Table 2 shows the intersection files between every two annotators. The total number of intersectional files is 302. They are used to compute the Inter Annotator Agreement (IAA) [4] to measure the understanding between the annotators and how well their annotations are consistent.

---

[1] https://www.sama.gov.sa/en-us/licenseentities/pages/licensedbanks.aspx [Date Accessed 2/4/2023]

[2] https://www.my.gov.sa/wps/portal/snp/agencies/!ut/p/z0/04_Sj9CPykssy0xPLMnMz0vMAfIjo8zivQIsTAwdDQz9_d29TAwCnQ1DjUy9wgwMgk31g1Pz9AuyHRUBX96rjw!!/ [Date Accessed 2/4/2023]

[3] https://chi.gov.sa/en/insurancecompanies/Pages/default.aspx [Date Accessed 2/4/2023]

**Table 2**

Annotation Distribution

| Intersection | Annotator1 | Annotator2 | Annotator3 |
|---|---|---|---|
| **Intersection 1 (103)** | Finance (1-52) E-Commerce (1-51) | | |
| **Intersection 2 (96)** | | Government (1-50) Healthcare (1-27) News (1-19) | |
| **Intersection 3 (103)** | Educational (1-23) Other (1-80) | | Educational (1-23) Other (1-80) |

Table 3 illustrates the results of IAA, Cohen kappa has been used to measure the agreement between each two annotators. The result shows almost perfect agreement between every two annotators, Annotator-1 & Annotator-2, Annotator-1 & Annotator-3, and Annotator-2 & Annotator-3, with an average of 95%, 96%, and 95%, respectively.

**Table 3**

Cohen Kappa Results

| Category | Annotators Agreement | | |
|---|---|---|---|
| | Annotator1 & Annotator2 | Annotator1 & Annotator3 | Annotator2 & Annotator3 |
| **Finance** | 0.96 | | |
| **E-commerce** | 0.94 | | |
| **Government** | | 0.92 | |
| **News** | | 0.98 | |
| **Healthcare** | | 0.98 | |
| **Educational** | | | 0.94 |
| **Other** | | | 0.97 |
| **Average** | 0.95 | 0.96 | 0.95 |

### 3) Data Cleaning and processing

In this step, we cleaned the text manually by extracting the pieces of information we needed in a text file. Furthermore, we eliminated the introduction and contact details. In addition, the copy/paste operation removed any hyperlinks from the text. After that, we combined all the privacy policy files in a single CSV file.

Next, in each policy line we performed the following.
- First, we removed the following Arabic numbers manually:
  - ( اولا، ثانيا، ثالثا، رابعا، خامسا، سادسا، سابعا، ثامنا، تاسعا، عاشرا، حادي عشر، ثاني عشر، ثالث عشر، رابع عشر)
  - (١, ٢, ٣, ٤, ٥, ٦, ٧, ٨, ٩)
- Then, using regular expressions in Python we removed Arabic Diacritics.

- In addition, we extracted English letters, symbols, and numbers. We also removed any blank lines, blank rows and multi spaces in each cell.

To have a homogeneous format we performed the following processes on the dataset.

(1) Normalization: we normalized the following:
  i. [آ] → [ا]
  ii. [ؤ] → [و]
  iii. [ى] → [ي]
  iv. [ة] → [ه]
  v. Tashkeel removal
  vi. Tatweel removal

(2) Tokenization: The words in each policy lines were separated into different tokens based on whitespaces using Whitespace Tokenizer.

We have investigated the use of stop words removal. However, we found that this removes the negation letters in Arabic which in turn changes the context of the privacy policy text and loses its significant meaning. Therefore, we decided not to use stop words removal in our work. As a result, we found out that the corpus contains 775,370 tokens, and the final size of the corpus equals to 8,353 KB.

## Ethics statements

The data used to compile the dataset do not pose any ethical concerns as they were collected from publically accessible websites.

## CRediT author statement

**Hend Al-Khalifa:** Conceptualization, Supervision, Project administration, Writing- Reviewing and Editing. **Malak Mashaabi:** Data curation, Investigation, Data Annotation, Original draft preparation. **Ghadi Al-Yahya**: Data curation, Investigation, Data Annotation, Original draft preparation. **Raghad Alnashwan:** Data curation, Investigation, Data Annotation, Original draft preparation.

## Declaration of interests

The authors declare that they have no known competing financial interests or personal relationships that could have appeared to influence the work reported in this paper.